\documentclass[a4, conference, final]{ieeeconf}

\makeatletter
\def\bstctlcite{\@ifnextchar[{\@bstctlcite}{\@bstctlcite[@auxout]}}
\def\@bstctlcite[#1]#2{\@bsphack
  \@for\@citeb:=#2\do{%
    \edef\@citeb{\expandafter\@firstofone\@citeb}%
    \if@filesw\immediate\write\csname #1\endcsname{\string\citation{\@citeb}}\fi}%
  \@esphack}
\makeatother

\usepackage[vlined, ruled, linesnumbered, commentsnumbered]{algorithm2e}
\usepackage{amsmath}
\usepackage{cite}
\usepackage{color}
\usepackage{xcolor}
\definecolor{chred}{rgb}{0.8,0,0}
\definecolor{chgray}{rgb}{0.5,0.5,0.5}
\usepackage{graphicx}
\usepackage{caption}
\usepackage{dblfloatfix}
\usepackage{subfigure}
\usepackage{eqlist}
\usepackage{txfonts}
\usepackage{url}
\usepackage{footmisc}
\usepackage{booktabs}

\usepackage{multirow}
\usepackage{threeparttable}
\usepackage{geometry}
\begin{document}
\newgeometry{top=25.4mm, bottom=19.1mm, left=19.1mm, right = 19.1mm} 
\bstctlcite{IEEEexample:BSTcontrol}
\IEEEoverridecommandlockouts
\title{Deep Learning Scooping Motion using Bilateral Teleoperations}

\author{Hitoe~Ochi, Weiwei~Wan, Yajue~Yang, Natsuki~Yamanobe, Jia~Pan,
		and Kensuke~Harada
\thanks{Hitoe~Ochi, Weiwei~Wan, and Kensuke~Harada are with Graduate School of
Engineering Science, Osaka University, Japan. Natsuki~Yamanobe,
Weiwei~Wan, and Kensuke~Harada are also affiliated with Natioanl
Institute of Advanced Industrial Science and Technology (AIST), Japan.
Yajue~Yang and Jia~Pan are with City University of Hongkong, China.
E-mail: {\tt\small wan@sys.es.osaka-u.ac.jp}}}
\maketitle

\begin{abstract}
We present bilateral teleoperation system for task learning and robot
motion generation. Our system includes a bilateral teleoperation platform and a
deep learning software. The deep learning software refers to human
demonstration using the bilateral teleoperation platform to collect visual
images and robotic encoder values. It leverages the datasets of images and
robotic encoder information to learn about the inter-modal correspondence
between visual images and robot motion. In detail, the deep learning software
uses a combination of Deep Convolutional Auto-Encoders (DCAE) over image regions, and Recurrent Neural
Network with Long Short-Term Memory units (LSTM-RNN) over robot motor angles, to
learn motion taught be human teleoperation.
The learnt models are used to predict new motion trajectories for
similar tasks. Experimental results show that our system has the
adaptivity to generate motion for similar scooping tasks. Detailed
analysis is performed based on failure cases of the experimental results.
Some insights about the cans and cannots of the system are summarized.
\end{abstract}



\section{Introduction}

\begin{figure}[!htbp]
	\centering
	\includegraphics[width=.47\textwidth]{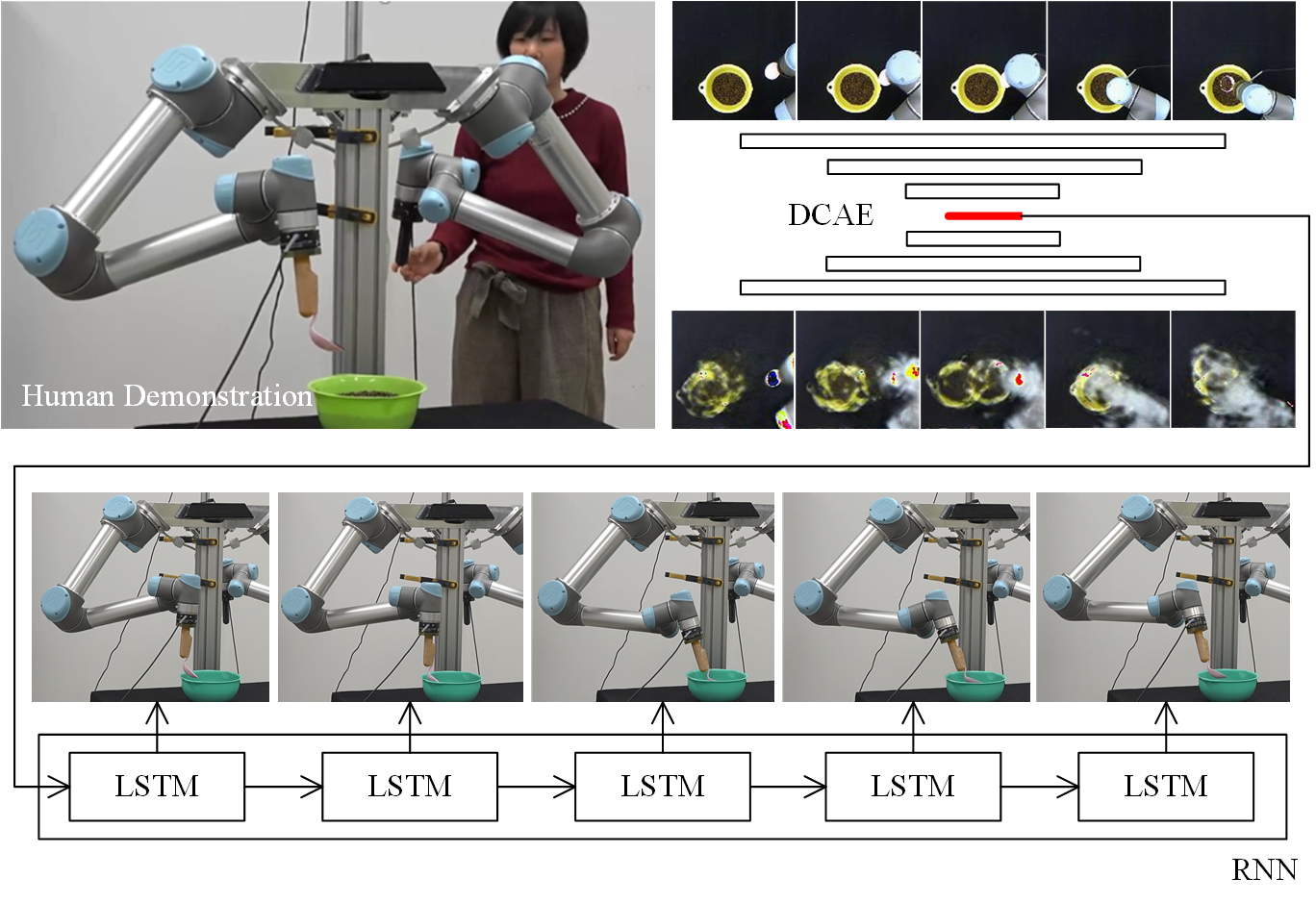}
 	\caption{The bilateral teleoperation system for task learning and robotic
	motion generation. The hardware platform of the system
	is a bi-lateral tele-operation systems composed of two same robot manipulators.
	The software of the system is a deep neural network made of a Deep
	Convolutional Auto-Encoder (DCAE) and a Recurrent Neural Network with Long Short-Term
    Memory units (LSTM-RNN). The deep learning models are trained by human
    demonstration, and used to generate new motion trajectories for similar tasks.}
	\label{teaser}
\end{figure}

Common household tasks require robots to act intelligently and adaptively in
various unstructured environment, which makes it difficult to model control
policies with explicit objectives and reward functions. One popular solution
\cite{argall2009survey}\cite{yang2017repeatable} is to circumvent the
difficulties by learning from demonstration (LfD).
LfD allows robots to learn skills from successful demonstrations performed by
manual teaching. In order to take advantage of LfD, we develop a system which
enables human operators to demonstrate with ease and enables robots to learn
dexterous manipulation skills with multi-modal sensed data. Fig.\ref{teaser} shows the system. The hardware platform of the system
is a bi-lateral tele-operation systems composed of two same robot manipulators. 
The software of the system is a deep neural network made of a Deep
Convolutional Auto-Encoder (DCAE) and a Recurrent Neural Network with Long Short-Term
Memory units (LSTM-RNN). The deep neural network leverages the datasets of images and
robotic encoder information to learn about the inter-modal correspondence
between visual images and robot motion, and the proposed system use the learnt
model to generate new motion trajectories for similar tasks.

Using the system, we can conduct experiments of robot learning different
basic behaviours using deep learning algorithms. Especially, we focus on a
scooping task which is common in home kitchen. We demonstrate that our system
has the adaptivity to predict motion for a broad range of scooping tasks.
Meanwhile, we examine the ability of deep learning algorithms with target
objects placed at different places and prepared in different conditions. We
carry out detailed analysis on the results and analyzed the reasons that
limited the ability of the proposed deep learning system.
We reach to a conclusion that although LfD using deep learning is applicable to
a wide range of objects, it still requires a large amount data to adapt to
large varieties. Mixed learning and planning is suggested to be a better
approach.

The paper is organized as follows. Section 2 reviews related work. Section 3
presents the entire LfD system including the demonstration method and the learning algorithm.
Section 4 explains how the robot perform a task after learning. Section 5
describes and analyzes experiment setups and results. Section 6 draws
conclusions and discusses possible methods to improve system performance.

\section{Related Work}

The learning method we used is DCAE and RNN. This section reviews their
origins and state-of-the-art applications.

Auto-encoders were initially introduced by Rumelhart et al.
\cite{rumelhart1985learning} to address the problem of unsupervised back
propagation. The input data was used as the teacher data to minimize
reconstruction errors \cite{olshausen1997sparse}.
Auto-encoders were embedded into deep neural network as DCAE to explore deep
features in \cite{hinton2006fast}\cite{salakhutdinov2009semantic}
\cite{bengio2007greedy}\cite{torralba2008small}, etc. DCAE helps to learn
multiple levels of representation of high dimensional data.

RNN is the feed-backward version of conventional feed forward neural network. It
allows the output of one neuron at time $t_i$ to be input of a neuron for
time $t_{i+1}$. RNN may date back to the Hopfield network
\cite{hopfield1982neural}. RNN is most suitable for learning and predicting
sequential data. Some successful applications of RNN include handwritting
recognition \cite{graves2009novel}, speech recognition \cite{graves2013speech},
visual tracking \cite{dequaire2017deep}, etc. RNN has advantages over
conventional mathematical models for sequential data like Hidden Markov Model (HMM) \cite{wan2007hybrid} in that it uses scalable historical
characters and is applicable to sequences of varying time lengths.

A variation of RNN is Multiple Timescale RNN (MTRNN), which is the multiple
timescale version of traditional RNN and was initially proposed by Yamashita and Tani
\cite{yamashita2008emergence} to learn motion primitives and predict new actions
by combining the learnt primitives. The MTRNN is composed of multiple Continuous
Recurrent Neural Network (CTRNN) layers that allow to have different timescale
activation speeds and thus enables scalability over time. Arie et al.
\cite{arie2009creating} Jeong et al. \cite {jeong2012neuro} are some other
studies that used MTRNN to generate robot motion.

RNN-based methods suffers from a vanishing gradient problem
\cite{hochreiter2001gradient}.
To overcome this problem, Hochereiter and Schmidhuber \cite{hochreiter1997long}
developed the Long Short Term Memory (LSTM) network.
The advantages of LSTM is it has an input gate, an output gate, and a forget
gate which allow the cells to store and access information over long periods of
time.

The recurrent neural network used by us in this paper is RNN with LSTM units. It
is proved that RNN with LSTM units are effective and scalable in long-range
sequence learning \cite{greff2017lstm}\footnote{There are some work like
\cite{yu2017continuous} that used MTRNN with LSTM units to enable multiple
timescale scalability.}. By introducing into each LSTM unit
a memory cell which can maintain its state over time, LSTM network is able to overcome the vanishing
gradient problem. LSTM is especially suitable for applications involving
long-term dependencies \cite{karpathy2015visualizing}.

Together with the DCAE, we build a system allowing predicting robot trajectories
for diverse tasks using vision systems. The system uses bilateral teleoperation
to collect data from human beings, like a general LfD system. It trains a DCAE
as well as a LSTM-RNN model, and use the model to learn robot motions to perform
similar tasks. We performed experiments by especially focusing on a scooping
task that is common in home kitchen.   
Several previous studies like
\cite{yang2017repeatable}\cite{mayer2008system}\cite{rahmatizadeh2017vision}\cite{liu2017imitation}
also studied learning to perform similar robotic tasks using deep learning models. Compared with them,
we not only demonstrate the generalization of deep models in robotic task
learning, but also carry out detailed analysis on the results and analyzed the
reasons that limited the ability of the proposed deep learning system. Readers
are encouraged to refer to the experiments and analysis section for details.

\section{The system for LfD using deep learning}

\subsection{The bilateral teleoperation platform}

Our LfD system utilizes bilateral teleoperation to allow human operators to
adaptively control the robot based on his control. Conventionally, teleoperation
was done in master-slave mode by using a joystick \cite{rahmatizadeh2017vision},
a haptic device \cite{abi2016visual}, or a virtual environment
\cite{liu2017imitation} as the master device.
Unlike the conventional methods, we use a robot manipulator as the master. As
Figure 2 shows, our system is composed of two identical robot systems comprising
a Universal Robot 1 arm at the same joint configuration and a force-torque sensor attached to
the arm's end-effector. The human operator drags the master at its end-effector
and the controller calculates 6 dimensional Cartesian velocity commands for
robots to follow the human operator's guidance. This dual-arm bilateral
teleoperation system provides similar operation spaces for the master and slave,
which makes it more convenient for the human operator to drag the master in a
natural manner. In addition, we install a Microsoft Kinect 1 above the slave
manipulator to capture depth and RGB images of the environment.

The bilateral teleoperation
platform provides the human operator a virtual sense of the contact force to
improve LfD \cite{hokayem2006bilateral}. While the human operator works on the
master manipulator, the slave robot senses a contact force with a force-torque sensor installed at its
wrist. A controller computes robot motions considering both the force
exerted by human beings and the force feedback from the force sensor.
Specifically, when the slave does not contact with the environment, both the
master and slave move following human motion. When the slave has contact
feedback, the master and slave react considering the impedance from
force feedback. The human operator, meanwhile, would feel the impedance from the
device he or she is working on (namely the master device) and react accordingly.

\subsection{The deep learning software}

\begin{figure}[!htbp]
	\centering
	\includegraphics[width=.47\textwidth]{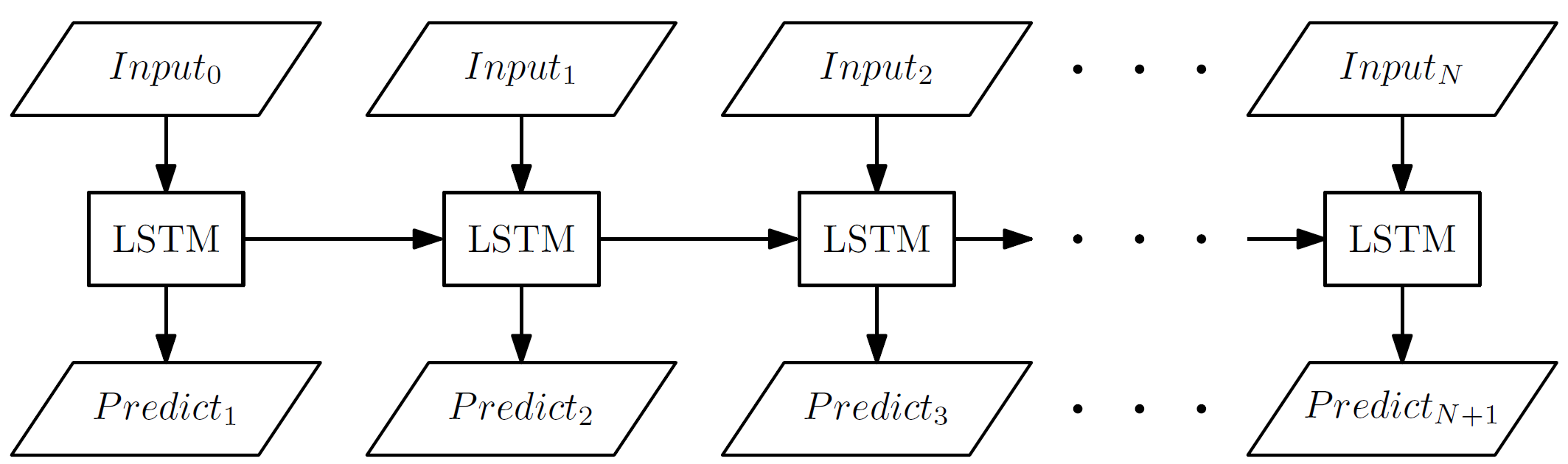}
 	\caption{LSTM-RNN: The subscripts in $Input_i$ and $Predict_i$ indicate the
 	time of inputs and for which predictions are made. An LSTM unit receives both
 	current input data and hidden states provided by previous LSTM units as inputs
 	to predict the next step.}
	\label{lstmrnn}
\end{figure}

LSTM-RNN supports both input and output sequences with variable
length, which means that one network may be suitable for varied tasks with
different length of time. Fig.\ref{lstmrnn} illustrates a LSTM recurrent network
which outputs prediction.

\begin{figure}[!htbp]
	\centering
	\includegraphics[width=.35\textwidth]{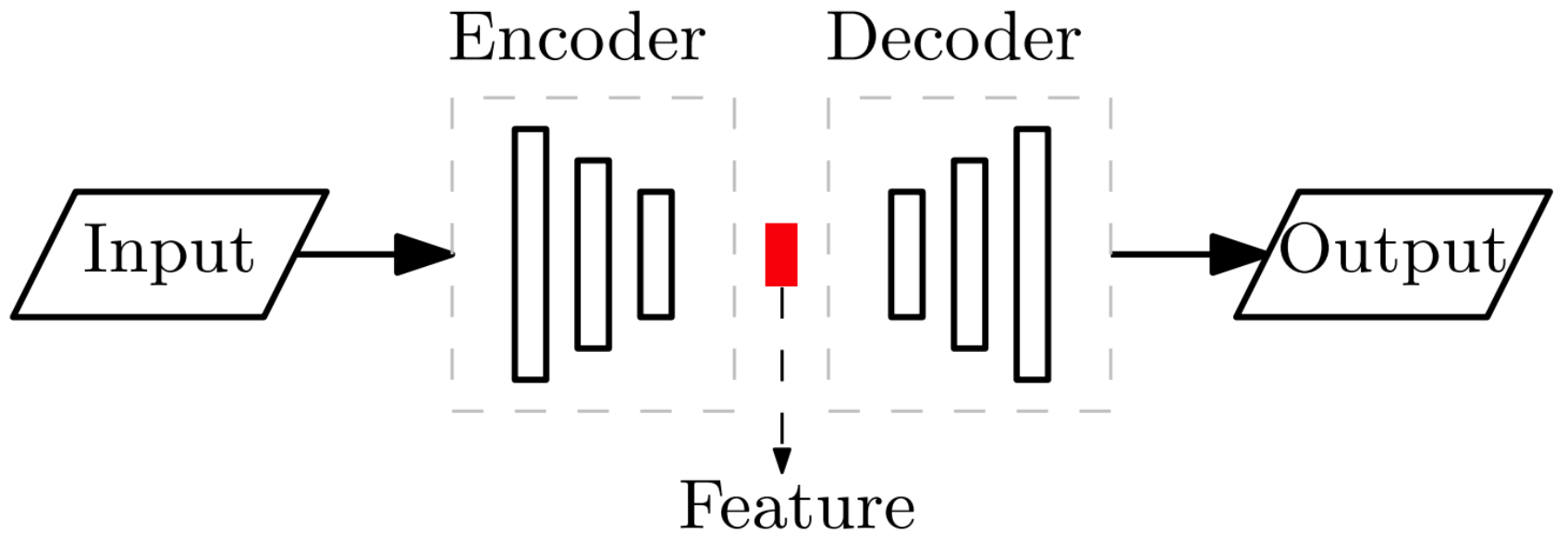}
 	\caption{DCAE encodes the input data with a encoder and reconstructs the data
	from the encoded values with a decoder. The output of DCAE (the intermediate
	layer) is the extracted data features.}
	\label{dcae}
\end{figure}

The data of our LfD system may include an image of the
environment, force/torque data sensed by a F/T sensor installed at the slave's
end-effector, robot joint positions, etc. These data has high dimensionality which makes computation
infeasible. To avoid the curse of dimensionality, we use DCAE to represent
the data with auto-selected features. DCAE encodes the input data
with a encoder and reconstructs the data from the encoded values with a decoder.
Both the encoder and decoder could be multi-layer convolutional networks
shown in Fig.\ref{dcae}. DCAE is able to properly encode complex data through
reconstruction, and extract data features and reduce data dimension.

\begin{figure}[!htbp]
	\centering
	\includegraphics[width=.49\textwidth]{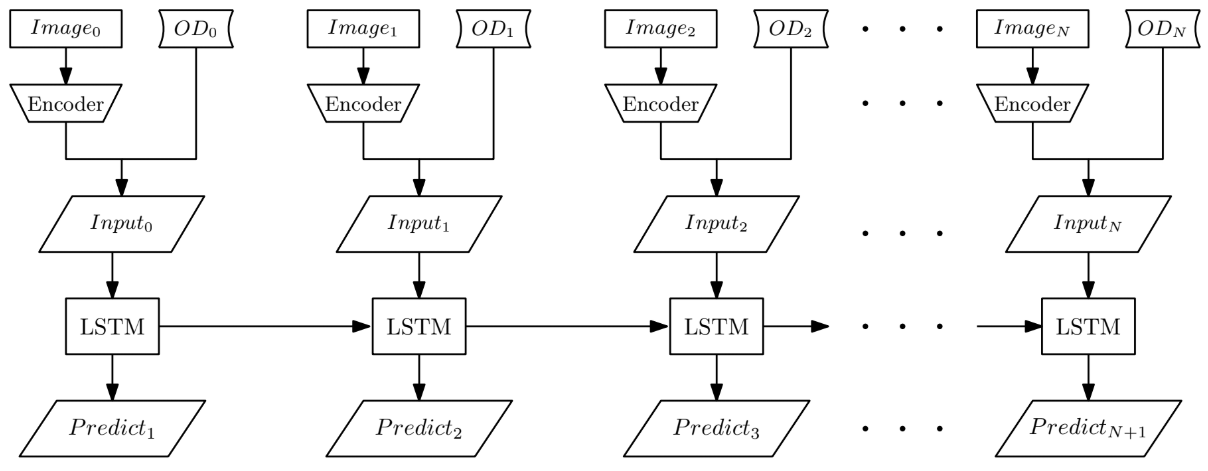}
 	\caption{The entire learning architecture. OD means Other Data. It could be
 	joint positions, force/torque values, etc. Although drawn as a single box, the
 	model may contain multiple LSTM layers.}
	\label{arch}
\end{figure}

The software of the LfD system is a deep learning architecture composed of DCAE
and LSTM-RNN. The LSTM-RNN model is fed with image features
computed by the encoder and other data such as joint positions, and predicts
a mixture of the next motion and surrounding situation.
The combination of DCAE and LSTM-RNN is shown in Fig.\ref{arch}.

\section{Learning and predicting motions}

\subsection{Data collection and training}

\begin{figure}[!htbp]
	\centering
	\includegraphics[width=.49\textwidth]{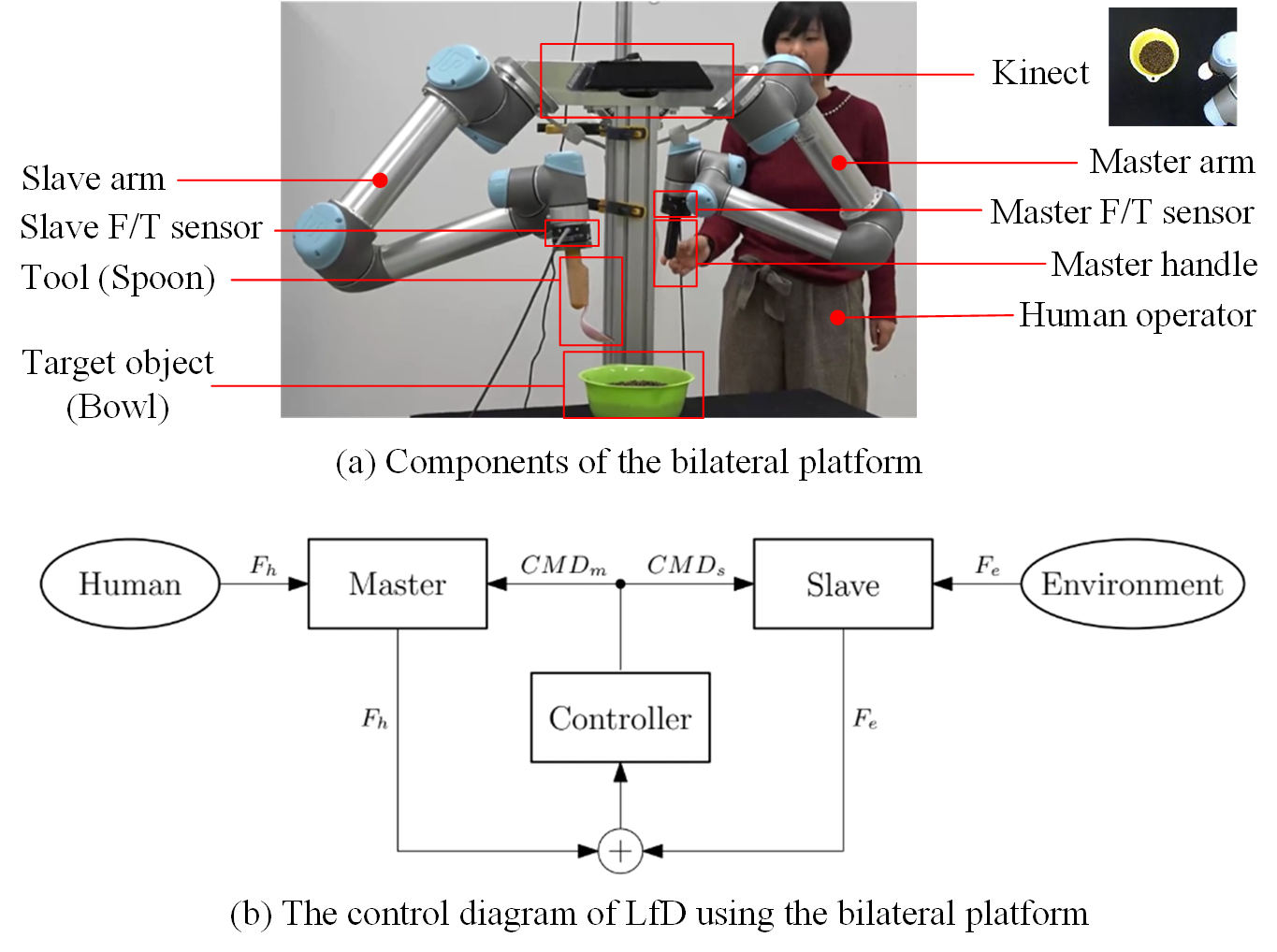}
 	\caption{Bilateral Teleoperation Diagram.}
	\label{bicontrol}
\end{figure}

\begin{figure}[!htbp]
	\centering
	\includegraphics[width=.49\textwidth]{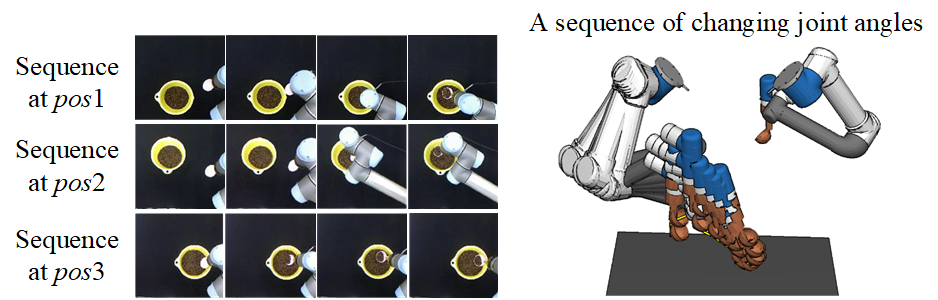}
 	\caption{The data used to train DCAE and LSTM-RNN. The left part shows the
 	snapshot sequences taken by the Kinect camera. They are used to train DCAE.
 	The right part shows the changing joint angles taught by human operators.
 	They are used to further train LSTM-RNN.}
	\label{sequences}
\end{figure}

The data used to train DCAE and LSTM-RNN is collected by bilateral
teleoperation. The components of the bilateral teleoperation platform and the
control diagram of LfD using the bilateral teleoperatoin platform are shown in
Fig.\ref{bicontrol}.
A human operator controls a master arm and performs a given task by considering
force feedback ($F_h\bigoplus F_e$ in the figure) from the slave side.
As the human operator moves the master arm, the Kinect camera installed at the
middle of the two arms take a sequence of snapshots as the training images for
DCAE. The motor encoders installed at each joint of the robot take a sequence of
6D joint angles as the training data of LSTM-RNN. The snapshots and changing
joint angles are shown in Fig.\ref{sequences}. Here, the left part shows three
sequences of snapshots. Each sequence is taken with the bowl placed at a different
position (denoted by $pos1$, $pos2$, and $pos3$ in the figure). The right part
shows a sequence of changing joint angles taught by the human operator.

\subsection{Generating robot motion} 

After training the DCAE and the LSTM-RNN, the models are used online to
generate robot motions for similar tasks. The trajectory generation
involves a real-time loop of three phases: (1) sensor data collection, (2) motion prediction, and
(3) execution. At each iteration, the current environment information and robot
state are collected and processed and then attached to the sequence of previous
data. The current robot state is fed to the pre-trained LSTM-RNN model to
predict next motions that a manipulator uses to take action. In order to
ensure computational efficiency, we keep each input sequence in a queue with
fixed length.

The process of training and prediction is shown in Fig.\ref{teaser}. Using the
pre-trained DCAE and LSTM-RNN, the system is able to generate
motion sequences to perform similar tasks.

\section{Experiments and analysis}

\begin{figure}[!htbp]
	\centering
	\includegraphics[width=.35\textwidth]{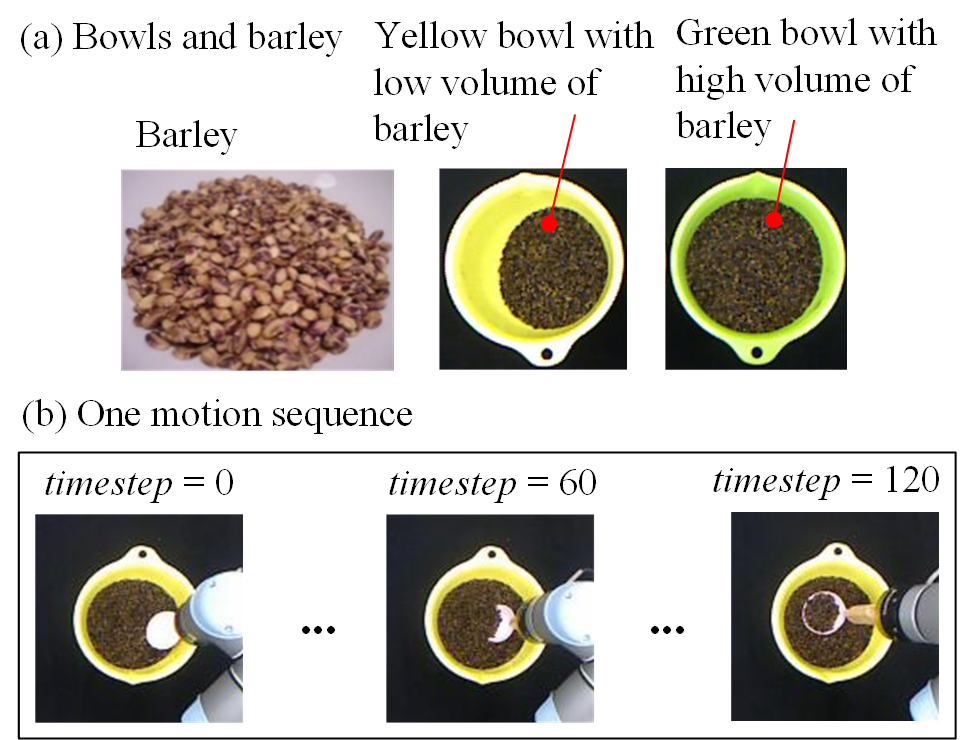}
 	\caption{(a) Two different bowls filled with different amount of barley are
 	used in experiments. In total, there are 2$\times$2=4 combinations. (b) One
 	sequence of scooping motion.}
	\label{expbowls}
\end{figure}

We use the developed system to learn scooping tasks. The goal of this task is to
scoop materials out from a bowl placed on a table in front of the robot (see
Fig.\ref{teaser}). Two different bowls filled with different amount of barley
are used in experiments. The two bowls include a yellow bowl and a green bowl. The volumes
of barley are set to ``high'' and ``low'' for variation. In total there
are 2$\times$2=4 combinations, namely \{``yellow'' bowl-``low'' barley,
``yellow'' bowl-``high'' barley, ``green'' bowl-``low'' barley, and
``green'' bowl-``high'' barley\}. Fig.\ref{expbowls}(a) shows the barley, the
bowls, and the different volume settings. During experiments, a human operator
performs teleoperated scooping as he/she senses the collision between the spoon
and the bowl after the spoon is inserted into the materials. Although used for
control, the F/T data is not fed into the learning system, which means the
control policy is learned only based on robot states and 2D images.

The images used to train DCAE are cropped by a 130$\times$130 window to lower
computational cost. The DCAE has
2 convolutional layers with filter sizes of 32 and 16, followed by 2
fully-connected layers of sizes 100 and 10. The decoder has exactly the same
structure. LeakyReLU activation function is used for all layers. Dropout is
applied afterwards to prevent over fitting. The computation is
performed on a Dell T5810 workstation with Nvidia GTX980 GPU.

\subsection{Experiment 1: Same position with RGB/Depth images}

In the first group of experiments, we place the bowl at the same position, and
test different bowls with different amounts of contents. In all, we collect
20 sequences of data with 5 for each bowl-content combination.
Fig.\ref{expbowls}(b) shows one sequence of the scooping motion. We use 19 of
the 20 sequences of data to train DCAE and LSTM-RNN and use the remaining one
group to test the performance.

Parameters of DCAE is as follows: Optimization function: Adam; Dropout rate:
0.4; Batch size: 32, Epoch: 50. We use both RGB images and Depth images to
train DCAE. The pre-trained models are named RGB-DCAE and Depth-DCAE
respectively.
Parameters of LSTM-RNN is: Optimization function: Adam; Batch size:
32; Iteration: 3000.

\begin{figure*}[!htbp]
	\centering
	\includegraphics[width=.99\textwidth]{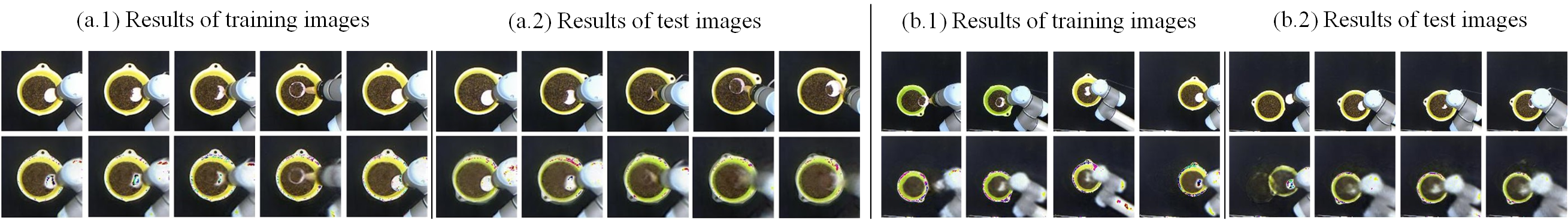}
 	\caption{The results of DCAE for experiment 1 and 2. (a.1-2) are the results
 	of training data and test data for experiment 1. (b.1-2) are the results of
 	training data and test data for experiment 2.}
	\label{dcaeresults}
\end{figure*}

The results of DCAE is shown in Fig.\ref{dcaeresults}(a). The trained model is
able to reconstruct the training data with high precision. Readers may compare
the first and second rows of Fig.\ref{dcaeresults}(a.1) for details. Meanwhile,
the trained model is able to reconstruct the test data with satisfying
performance. Readers may compare the first and second row of
Fig.\ref{dcaeresults}(a.2) to see the difference. Although there are noises on
the second rows, they are acceptable.

The results of LSTM-RNN show that the robot is able to perform scooping for
similar tasks given the RGB-DCAE. However, it cannot precisely differ ``high''
and ``low'' volumes. The results of LSTM-RNN using Depth-DCAE is unstable. We
failed to spot a successful execution.
The reason depth data is unstable is probably due to the low resolution of
Kinect's depth sensor. The vision system cannot differ if the spoon is at a
pre-scooping state or post-scooping state, which makes the robot hard to predict
next motions.

\subsection{Experiment 2: Different positions}

In the second group of experiments, we place the bowl at different positions to
further examine the generalization ability of the trained models. 

Similar to experiment 1, we use bowls with two different colors (``yellow''
and ``green''), and used two different volumes of contents (``high'' and ``low'').
The bowls were placed at 7 different positions. At each position, we collect
20 sequences of data with 5 for each bowl-barley combination. In total, we
collect 140 sequences of data. 139 of the 140 sequences of data are used to
train DCAE and LSTM-RNN. The remaining 1 sequence are used for testing. The
parameters of DCAE and LSTM-RNN are the same as experiment 1.

The results of DCAE are shown in Fig.\ref{dcaeresults}(b). The trained model is
able to reconstruct the training data with high precision. Readers may compare
the first and second rows of Fig.\ref{dcaeresults}(b.1) for details. In
contrast, the reconstructed images show significant difference for the test
data. It failed to reconstruct the test data. Readers may compare the first and
second row of Fig.\ref{dcaeresults}(b.2) to see the difference. Especially for
the first column of Fig.\ref{dcaeresults}(b.2), the bowl is wrongly considered
to be at a totally different position.

The LSTM-RNN model is not able to generate scooping motion for either the
training data or the test data. The motion is randomly changing from time to
time. It doesn't follow any pre-taught sequences. The reason is probably the bad
reconstruction performance of DCAE. The system failed to correctly find the
positions of the bowls using the encoded features. Based on the analysis,
we increase the training data of DCAE in Experiment 3 to
improve its reconstruction.

\subsection{Experiment 3: Increasing the training data of DCAE}

The third group of experiments has exactly the same scenario settings and
parameter settings as Experiment 2, except that we use planning algorithms to
generate scooping motion and collect more scooping images.

\begin{figure}[!htbp]
	\centering
	\includegraphics[width=.43\textwidth]{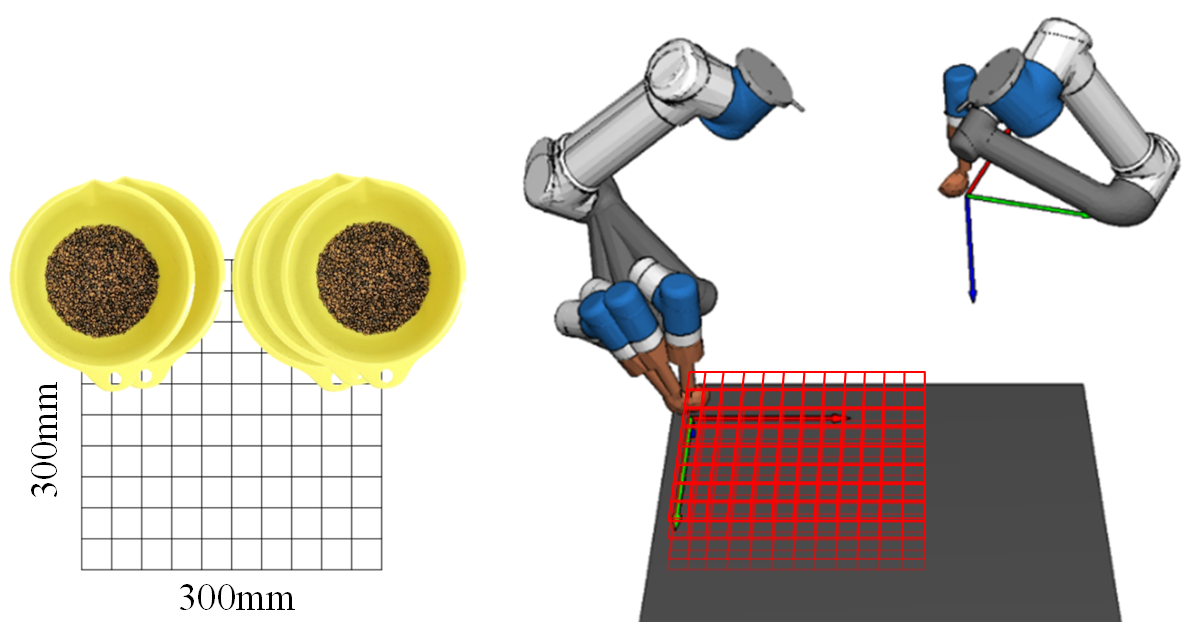}
 	\caption{Increase the training data of DCAE by automatically generating
 	motions across a 10$\times$10 grids. In all, 100$\times$45$\times$3=13500
 	extra training images were generated. Here, ``100'' indicates the 100 grid
 	positions. At each grid, 45 arm positions and 3 arm rotation angles are
 	sampled.}
	\label{sample}
\end{figure}

The new scooping images are collected following the work flow shows in
Fig.\ref{sample}. We divide the work space into around 100 grids, place bowl at
these places, and sample arm boatswains and orientations at each of the grid. In
total, we additionally generate 100$\times$45$\times$3=13500 (12726 exactly)
extra training images to train DCAE. Here, ``100'' indicates the 100 grid
positions. ``45'' and ``3'' indicate the 45 arm positions and 3 arm rotation
angles sampled at each grid.

\begin{figure*}[!htbp]
	\centering
	\includegraphics[width=.99\textwidth]{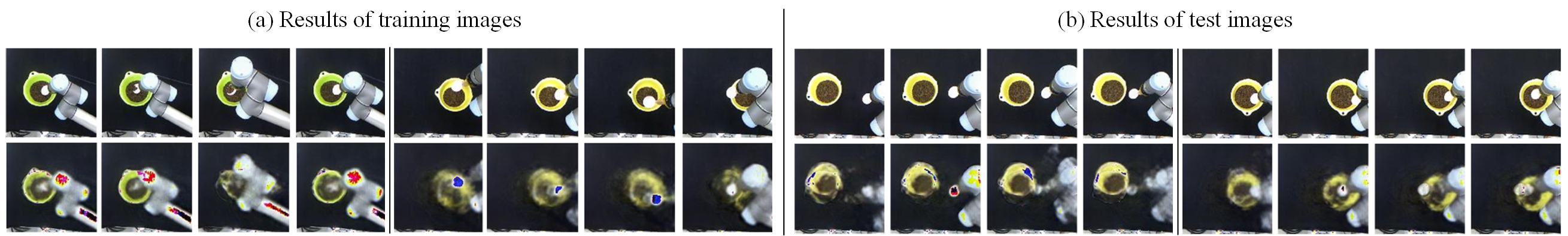}
 	\caption{The results of DCAE for experiment 1 and 2. (a.1-2) are the results
 	of training data and test data for experiment 1. (b.1-2) are the results of
 	training data and test data for experiment 2.}
	\label{dcaeresults2}
\end{figure*}

The DCAE model is trained with the 140 sequences of data in experiment 2 (that
is 140$\times$120=16800 images, 17714 exactly), together with the 13500
extra images collected using the planner. The parameters of DCAE are exactly the same as Experiment 1
and 2. The results of DCAE is shown in Fig.\ref{dcaeresults2}. Compared with
experiment 2, DCAE is more stable. It is able to reconstruct both the training
images and the test images with satisfying performance, although the
reconstructed spoon position in the sixth and seventh columns of
Fig.\ref{dcaeresults2}(b) have relatively large offsets from the original image.

The trained DCAE model is used together with LSTM-RNN to predict motions. The
LSTM-RNN model is trained using different data to compare the performance. The
results are shown in Table.\ref{conf0}. Here, $A1$-$A7r\_s1$ indicates the data
used to train DCAE and LSTM-RNN. The left side of ``\_'' shows the data used to train
DCAE. $A1$-$A7$ means all the sequences collected at the seven bowl positions in
experiments 2 are used. $r$ means the additional data collected in experiment 3
is used. The right side of ``\_'' shows the data used to train LSTM-RNN. $s1$
means only the sequences at bowl position $s1$ are used to train LSTM-RNN.
$s1s4$ means both the sequences at bowl position $s1$ and position $s4$ are used
to train LSTM-RNN. 

The results show that the DCAE trained in experiment 3 is able to predict motion
for bowls at the same positions. For example, (row position $s1$, column
$A1$-$A7r\_s1$) is $\bigcirc$, (row position $s1$, column $A1$-$A7r\_s1s4$) is also $\bigcirc$. The
result, however, is unstable. For example, (row position $s2$, column
$A1$-$A7r\_s1s4$) is $\times$, (row position $s4$, column
$A1$-$A7r\_s4$) is also $\times$. The last three columns of the table shows
previous results: The $A1\_s1$ column and $A4\_s4$ column correspond to the
results of experiment 1. The $A1A4\_s1s4$ column correspond to the result of experiment 2.

\begin{table*}[!htbp]
\centering
\renewcommand{\arraystretch}{1.2}
\caption{\label{conf0}Results of scooping using DCAE and LSTM-RNN}
\begin{tabular}{ccccccc}
\toprule
& A1-A7r\_s1 & A1-A7r\_s1s4 &  A1-A7r\_s4 &  A1\_s1 & A4\_s4 & A1A4\_s1s4\\
 \midrule
 \midrule
 position $s1$ & $\bigcirc$ & $\bigcirc$ & - & $\bigcirc$ & - & $\times$\\
 \midrule
 position $s4$ & - & $\times$ & $\times$ & - & $\bigcirc$ & $\times$\\
\midrule
\bottomrule
\end{tabular}
\end{table*}

Results of the three experiments show that the proposed model heavily depends
on training data. It can predict motion for different objects at the same
positions, but is not able to adapt to objects at different positions. The
small amount of training data is an important problem impairing the
generalization of the trained models to different bowl positions. The experimental results tell
us that a small amount of training data leads to bad results. A large amount of
training data shows good prediction.

\section{Conclusions and future work}

This paper presented a bilateral teleoperation system for task learning and
robotic motion generation. It trained DCAE and LSTM-RNN to learn scooping
motion using data collected by human demonstration on the bilateral
teleoperation system. The results showed the data collected using the bilateral
teleoperation system was suitable for training deep learning models. The trained
model was able to predict scooping motion for different objects at the same
positions, showing some ability of generalization. The results also showed that
the amount of data was an important issue that affect training good deep
learning models.

One way to improve performance is to increase training data. However,
increasing training data is not trivial for LfD applications since they require
human operators to repeatedly work on teaching devices. Another method is to
use a mixed learning and planning model. Practitioners may use planning to
collect data and use learning to generalize the planned results. The mixed
method is our future direction.

\section*{Acknowledgment}
The paper is based on results obtained from a project commissioned by the
New Energy and Industrial Technology Development Organization (NEDO).

\bibliographystyle{IEEEtran}
\bibliography{reference}

\end{document}